\documentclass[10pt,twocolumn,letterpaper]{article}

\usepackage{iccv}
\usepackage{times}
\usepackage{epsfig}
\usepackage{graphicx}
\usepackage{amsmath}
\usepackage{amssymb}
\usepackage[accsupp]{axessibility}

\usepackage{color}
\definecolor{darkred}{rgb}{0.7,0.1,0.1}
\definecolor{red}{rgb}{0.9,0.05,0.05}
\definecolor{darkgreen}{rgb}{0.1,0.7,0.1}
\definecolor{cyan}{rgb}{0.7,0.0,0.7}
\definecolor{dblue}{rgb}{0.2,0.2,0.8}
\definecolor{maroon}{rgb}{0.76,.13,.28}
\definecolor{burntorange}{rgb}{0.81,.33,0}
\definecolor{blue_r}{rgb}{0.4392,0.1882,0.6274}
\definecolor{blue_s}{rgb}{0.0,0.44,0.75}
\definecolor{blue_a}{rgb}{0.0,0.69,0.941}

\ifdefined\ShowNotes
  \newcommand{\colornote}[3]{{\color{#1}\bf{#2: #3}\normalfont}}
\else
  \newcommand{\colornote}[3]{}
\fi

\usepackage[breaklinks=true,bookmarks=false]{hyperref}

\iccvfinalcopy 

\def\iccvPaperID{4962} 

\ificcvfinal\fi

\begin{document}

\title{A Game of Bundle Adjustment - Learning Efficient Convergence}

\author{Amir Belder \thanks{Equal contribution}
\\
Technion and Reality Labs, Meta inc.\\
{\tt\small amirbelder@campus.technion.ac.il}
\and
 Refael Vivanti* \\ 
Reality Labs, Meta inc.\\
{\tt\small refaelv@fb.com}
\and
Ayellet Tal\\
Technion and Cornel-Tech\\
{\tt\small ayellet@ee.technion.ac.il}
}

\maketitle
\ificcvfinal\thispagestyle{empty}\fi

\begin{abstract}
Bundle adjustment is the common way to solve localization and mapping.
It is an iterative process in which a system of non-linear equations is solved using two optimization methods, weighted by a damping factor.
In the classic approach, the latter is chosen heuristically by the Levenberg-Marquardt algorithm on each iteration. 
This might take many iterations, making the process computationally expensive, which might be harmful to real-time applications.
We propose to replace this heuristic by viewing the problem in a holistic manner, as a game, and formulating it as a reinforcement-learning task.
We set an environment which solves the non-linear equations and train an agent to choose the damping factor in a learned manner.
We demonstrate that our approach considerably reduces the number of iterations required to reach 
the bundle adjustment's convergence, on both synthetic and real-life scenarios.  
We show that this reduction benefits the classic approach and can be integrated  with other bundle adjustment acceleration methods. 

\end{abstract}


\section{Introduction}
\label{sec:intro}

\begin{figure}[t]
\centering
\begin{tabular}{c}
\includegraphics[width=0.43\textwidth]{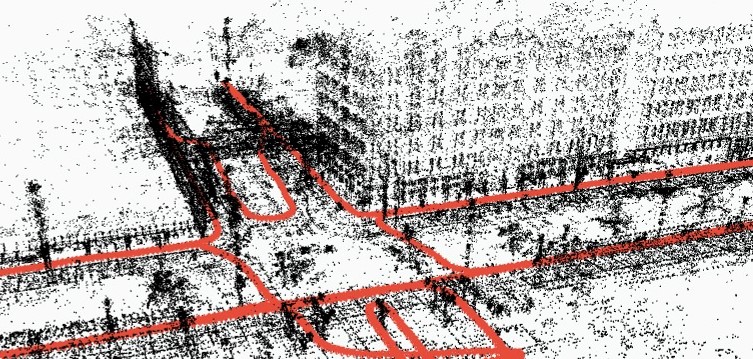} \\

\end{tabular}
\caption{
Given a series of 2D images taken by a camera from different positions, the iterative {\em Bundle Adjustment (BA)} process evaluates the 3D locations of the objects in the images (in black) and the camera's poses, as seen in the {\color{red}{red} trajectory}.
We propose a method to accelerate the process by reducing the number iterations required for the solving.  
}
\label{fig:teaser}
\end{figure}

{\em Simultaneous Localization And Mapping (SLAM)} is successfully used in numerous fields, including computer vision~\cite{triggs1999bundle}, augmented reality~\cite{agarwal2010bundle, ni2007out, zhou2020stochastic} and autonomous driving~\cite{ni2007out, zhou2020stochastic, ortiz2020bundle}.
Its input is a series of 2D images of a scene taken by a single camera from different viewpoints, from which a set of 2D {\em matches} 
are extracted. 
The goal is to estimate the objects' 3D 
locations, 
and the camera's poses
(locations and angles) throughout the capturing 
according to the 2D matches. 
See Fig.\ref{fig:teaser}
where the 3D locations appear in black and the camera's 
poses form the trajectory in {\color{red}{red}}.
{\em Structure From Motion (SFM)} is a similar process where the images are taken by several cameras~\cite{zhu2018very, zhu2017parallel, ni2007out, fang2019merge}. 

SLAM is 
commonly solved using the iterative {\em Bundle Adjustment (BA)} process~\cite{foresee1997gauss, more1978levenberg}.
In fact, BA occupies roughly $60\%$–$80\%$ of the execution time needed for the mapping~\cite{tanaka2021learning}.
On each iteration the 3D locations and camera poses are first evaluated by a combination of two optimization methods: 
{\em Gradient descend (GD)} and {\em Gauss-Newton (GN)}, 
which are weighted according to a {\em damping factor}, termed {\em $\lambda$}. 
Then, the evaluated locations are projected into 2D according to the evaluated poses.
The stopping
criterion (convergence) of this iterative process is usually met when the difference between the evaluated 2D projections and the initially extracted 2D matches (termed {\em projection error}) is lower than a certain threshold. 
Due to computational constraints, if convergence is not achieved within a fixed number of iterations, the process is stopped.


Two main factors influence the execution time:
(1)~the duration of a single iteration, which is mainly affected by the Hessian's calculation that GN entails;
(2)~the required number of iterations to reach convergence, caused by inefficient choosing of~$\lambda$.
Some previous BA acceleration methods focus on the first factor and reduce the duration of each iteration, by suggesting efficient ways to calculate and invert the sparse Hessian~\cite{zhou2020stochastic,  huang2021deeplm}.
The focus of this paper is on the second factor---decreasing the number of iterations.

In the classic approach, the value of $\lambda$ is determined heuristically by the
{\em Levenberg-Marquardt (LM)} 
algorithm~\cite{more1978levenberg} on each iteration.
It may change only by one of two specific constant factors between consecutive iterations.
This limits the ability to efficiently change the optimization scheme between GD and GN, even when it can be beneficial. 
%
%
We propose to address this problem differently.

Our key idea is to {\em learn a dynamic} value of $\lambda$.
As the choice of $\lambda$'s value on each iteration may influence the solving for {\em several} iterations,
we propose to view the process in a new light.
Differently from previous approaches, we view the BA process in a holistic manner 
as a game.
%
We show how a simple {\em Reinforcement Learning  (RL)} framework suffices
%
to achieve a solution that upholds a dynamic and efficient weighting of GD and GN, which is determined by $\lambda$.
Briefly, RL tasks are defined by an {\em environment} and an {\em agent}. 
The agent learns to preform actions according to the environment's response to these actions (at the form of {\em rewards}).
The agent aims to maximize the sum of the expected rewards, which is the key to handling delayed and sparse rewards like the BA's single and delayed convergence. 
In our case, the environment solves the BA problem and its step performs a single BA iteration. 
As we aim at a learned $\lambda$, we chose to represent the value of $\lambda$ as the agent's action. 
The reward is positive only on the iteration convergence is achieved and is negative 
otherwise.
Therefore, in every iteration convergence is not achieved the agent gets a negative reward as a "fine".  
Since the agent aims at maximizing the sum of the expected rewards, it is encouraged to find a valid solution (reach convergence) within as few iterations as possible.

Our method is shown to reduce the number of iterations required to achieve the BA convergence by a factor of $3$-$5$ on both KITTI~\cite{geiger2013vision} 
and BAL~\cite{agarwal2010bundle} benchmarks. 
Furthermore, our approach is likely to impact common real-life BA problems, whose solving may require much time due to their large size.
In addition, we demonstrate that our agent could be trained in a time-efficient manner on small synthetic scenes of randomly chosen locations and camera's poses, 
and still accelerate the solving of real-life scenarios.
Finally, our approach may be integrated and added to previous works that focus on reducing the time of each iteration~\cite{zhou2020stochastic,  huang2021deeplm}, but might be less beneficial to other methods such as~\cite{ortiz2020bundle} that suggest  hardware-based  BA optimizations.

Hence our work makes the following contributions:
\begin{enumerate}
    \item 
    We propose a general and unified approach that learns the ideal value of $\lambda$.
    It can be integrated within other BA acceleration methods. 
    \item
    We propose a network that utilizes this approach using Reinforcement Learning.
    We show that it achieves a significant reduction in the number of iterations and running time. On the KITTI benchmark for instance, a $1/5$ of the iterations were required, which led to an overall speedup of $~3$. 
\end{enumerate}
\section{Related Work}
\noindent
{\bf{Bundle Adjustment (BA).}} This is a known method to address {\em Simultaneous Localization And Mapping (SLAM)}~\cite{agarwal2010bundle,ni2007out,triggs1999bundle,wu2011multicore,zhou2020stochastic} problems.
Given a set of 2D key-points (matches), 
BA~\cite{foresee1997gauss, more1978levenberg} aims to solve a system of non-linear equations to evaluate the 3D locations and camera poses
according to those matches. 
Due to the non-linear nature of the equations BA is solved iteratively.
On each iteration a {\em Reduced Camera System}~\cite{jeong2011pushing, lourakis2009sba} 
is solved 
by two optimization methods 
that 
are weighted by a damping factor, $\lambda$.
$\lambda$'s value is determined by the {\em Levenberg-Marquardt (LM)}~\cite{more1978levenberg} algorithm's heuristic on each iteration as follows: $\lambda$ is multiplied by $1/2$ if the current iteration's estimation error is larger than that of the previous iteration, or by $2$ otherwise. 
\vspace{0.1in}
\noindent
{\bf{Bundle Adjustment Acceleration.}} 
As BA is a fundamental problem in various fields and a main efficiency bottleneck for many real-time applications, several works that accelerate it were introduced
~\cite{huang2021deeplm, ortiz2020bundle, tanaka2021learning, zhou2020stochastic, Demmel_2021_CVPR, Demmel_2021_ICCV}.
Each work faces the acceleration 
challenge differently. 
Tanaka et al.~\cite{tanaka2021learning} try to replace the BA process entirely by splitting the solving into smaller ("local") parts, and solve each local part using a {\em Neural Network (NN)}.
ang and Tan \cite{tang2018banet} also use a NN to solve BA, and learn $\lambda$'s value for a constant number of iterations ($5$).
Von Stumberg et al.~\cite{DBLP:journals/corr/abs-1904-11932} introduce a Gauss-Newton loss to learn SLAM solving.
Oritz et al.~\cite{ortiz2020bundle} replace LM with {\em Gaussian Belief Propagation (GBP)} which requires a separate damping factor for each key-point, and use {\em Intelligence Processing Unit (IPU)} hardware to improve parallelism capabilities.
In~\cite{Demmel_2021_CVPR, Demmel_2021_ICCV} Demmel et al. utilize fixed point approximations to accelerate the solving.
Other methods focus on accelerating the time of a single iteration.
Both Clark et al.~\cite{DBLP:journals/corr/abs-1809-02966} and Zhou et al.~\cite{zhou2020stochastic} use a NN to calculate the Jacobian matrix. 
For instance, ~\cite{zhou2020stochastic} split the BA into smaller problems via clustering. 
Huang et al.~\cite{huang2021deeplm} use domain decomposition to split the solving into smaller clusters. 
~\cite{DBLP:journals/corr/AmosK17, DBLP:journals/corr/abs-1910-01727} may also be used to optimize BA's performance. 
Unlike past works, we focus on 
reducing the sheer number of iterations of the LM based BA solving.
Our iteration reduction could therefore benefit several previous approaches and be added to them. 

\vspace{0.1in}
\noindent
{\bf{Reinforcement Learning (RL).}} 
This is a growing field of research in machine leaning that has been used for many applications~\cite{li2017deep, sutton2018reinforcement, DBLP:journals/corr/abs-2104-05565, DBLP:journals/corr/abs-2108-11510, mazyavkina2021reinforcement, DBLP:journals/corr/abs-1906-11527,ni2007out}.
RL problems are commonly represented by an agent and an environment.
Each time the agent preforms an action ($a$), the environment responds by preforming a step according to that action and returns an observation (state $s$) and a reward ($r$).
RL problems are defined by their actions, states and rewards that can be either discrete or continuous and of any dimension. 
The agent chooses its actions according to a stochastic policy $\pi$, which determines the probability to choose each action in the action space.
The state provides information about the environment, like the estimation error in our case, while the reward  encourages the agent to reach convergence.
Value functions~($v$) evaluate the sum of expected future rewards, and are evaluated according to a specific policy, i.e. $v_{\pi}$.

RL is used in various fields including games~\cite{sutton2018reinforcement}, robotics~\cite{kober2013reinforcement}, {\em Natural Language Processing (NLP)}~\cite{ DBLP:journals/corr/abs-2104-05565} and {\em Computer Vision (CV)}~\cite{DBLP:journals/corr/abs-2108-11510}.
It is also used for hyper-parameter tuning of both classic~\cite{mazyavkina2021reinforcement} and {\em Deep Leaning (DL)}~\cite{DBLP:journals/corr/abs-1906-11527}  optimization methods.
This work is among the first to utilize {\em Deep Reinforcement Learning}
to choose $\lambda$'s value to accelerate the BA's optimization problem solving.

\vspace{0.1in}
\noindent
{\bf{Soft Actor Critic (SAC).}} This is a RL framework that aims at augmenting the standard RL maximum reward objective with an entropy maximization term, which leads to a substantial improvement in exploration~\cite{haarnoja2018soft, christodoulou2019soft}.
In their work, Haarnoja et al.~\cite{haarnoja2018soft} show that SAC achieves fast and stable convergence on various RL tasks.
We chose SAC as our RL framework as it is stable and suites continuous state and action spaces, like in our BA problem.
Furthermore, the large number of parameters that need to be updated and estimated on each BA iteration results in many possible solutions. 
Such a large and complex problem could greatly benefit from SAC's extensive and sophisticated exploration.
\section{Method}

Given a series of 2D images taken by a single camera, the {\em Bundle Adjustment (BA)} iterative optimization process 
aims at evaluating the camera's poses and objects' 3D locations.
Two optimization methods are used for the evaluation on each iteration: 
{\em Gradient Descend (GD)} and {\em Gauss-Newton (GN)}, that 
are weighted by a damping factor, $\lambda$. 

Although both GD and GN advance in the direction of the gradient, they differ in nature.
Generally speaking, the GN takes a bigger step than the GD and is well suited to explore parabolic functions. 
But, 
if the local function is not parabolic in nature, the big GN step might divert the solution away from the true minima.
In such cases the GD is more effective.
But relying on the small GD step alone could lead to a slow and inefficient solving.
Therefore, efficient BA solving requires efficient weighing between GD and GN, which is determined by $\lambda$.
Recall that in the classic approach, $\lambda$'s value is set by the {\em Levenberg-Marquardt (LM)} algorithm and may change by one of two constant factors between consecutive iterations. 
This may result in inefficient weighting of GD and GN and
consequently in a large number of iterations until convergence is achieved.

Our key idea is therefore to learn a {\em dynamic} value of $\lambda$, to dynamically weight the two optimization methods in an efficient manner. 
This is not straight-forward as the BA's convergence (or failure) is achieved only once at the very end of the solving process,
and since each choice of $\lambda$ may affect the solving for several iterations.
Therefore, as we aim to reduce the total number of iterations required to reach convergence,
the learning of the ideal value of $\lambda$ requires viewing the solving process as a whole.
%

Hence, differently from previous methods, we propose to view the BA solving process in a new and holistic manner as a game.
We may draw an analogy to a chess game, where victory (convergence) is achieved only once at the very end of the game, while each turn (iteration) may go better or worse (estimation error) and the choosing of $\lambda$'s value is analogous to choosing a chess piece and moving it.

Fortunately, {\em Reinforcement Learning (RL)} methods are designed to handle continuous processes, such as the BA's convergence, while producing long-term decisions.
This is done by viewing each process in holistic manner, that enables handling
 sparse and delayed rewards. 
Hence, RL could be harnessed to learn the optimal value of $\lambda$. 
Thus, we formulate the BA problem in RL terms by defining states, actions and rewards.
As our method is not limited by two constant factors between iterations, it enables a dynamic and efficient weighting of GD and GN along the solving. 
This may reduce the number of iterations required for convergence and result in a more time-efficient solving process.

\begin{figure}[tb]
\centering
\begin{tabular}{c}
\includegraphics[width=0.48\textwidth]{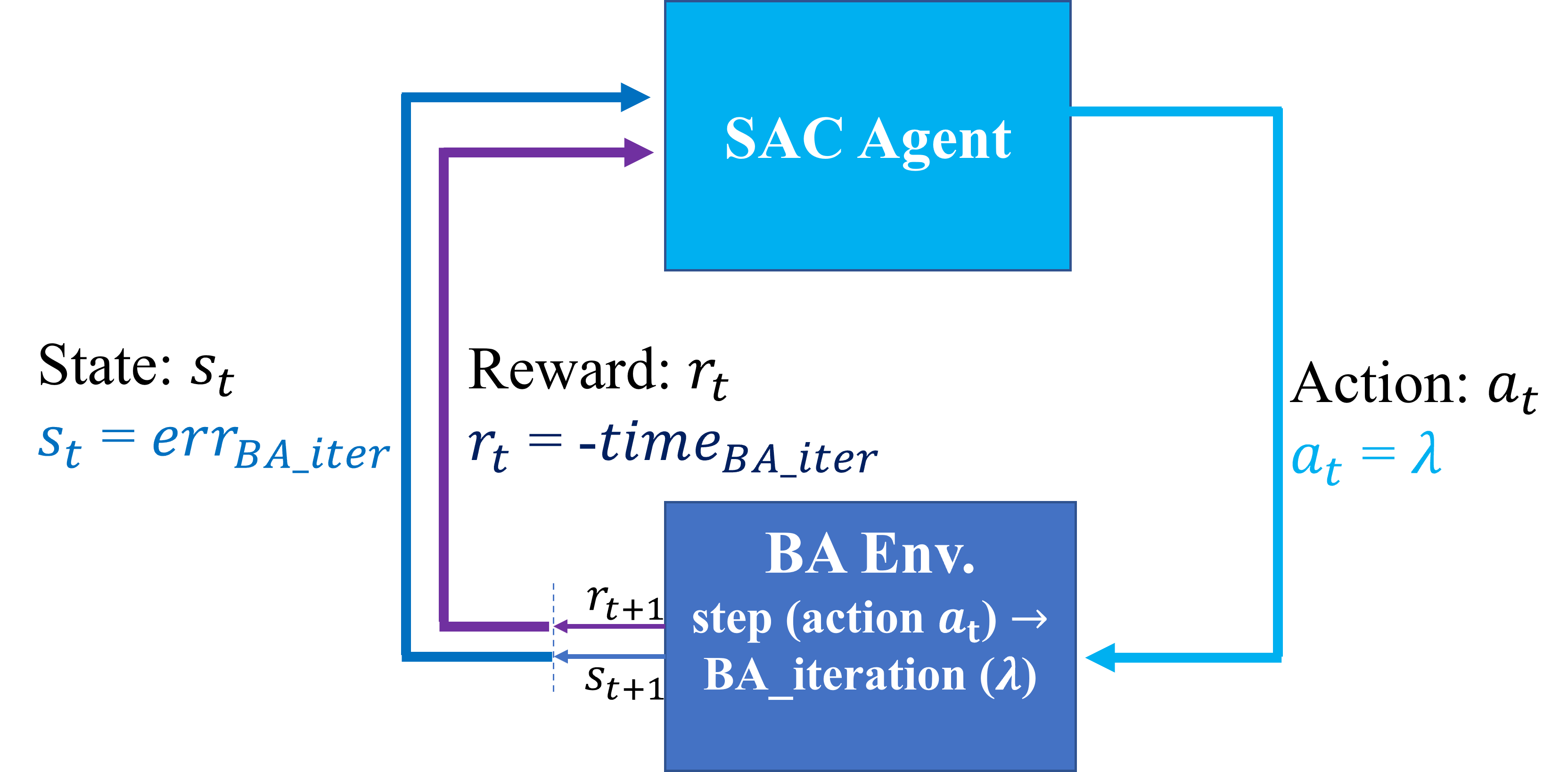}
\end{tabular}
\caption{
{\bf BA problem in RL terms.} The SAC agent chooses $\lambda$'s value as its  {\color{blue_a}{action ($a$)}}, 
and then the environment preforms a single BA iteration (step), where GD and GN are weighted according to~$\lambda$.
The environment responds with:
1. a  {\color{blue_s}{state ($s$})} which represents the estimation error of the BA's iteration;
2. a  {\color{blue_r}{reward ($r$)}} that represents the iteration's duration as a negative (in seconds), except for the iteration convergence is met, where {\color{blue_r}{$r$}} serves as a positive convergence bonus.
As the agent aims at maximizing the sum of expected rewards, it is encouraged to choose $\lambda$ in a manner that reduces the number of solving iterations.}
 \label{fig:rl}
 \end{figure}

We use the {\em Soft Actor Critic (SAC)} RL framework, as it is stable and adjusted to continuous state and action spaces like our problem entails, and for its extensive exploration which is beneficial in our highly complex and multi-variable problem~\cite{haarnoja2018soft}.
Our method consists of two main parts.
The first is an environment which solves a single BA iteration on each step and provides a state and a reward according to it.
The second is the SAC agent that predicts the value of $\lambda$ as its action; see Fig.~\ref{fig:rl}.
We elaborate on each part hereafter. 

%
\vspace{0.1in}
\noindent
{\bf Environment.} 
The environment solves a BA problem.
During its initialization, the environment gets a set of 2D matches, representing the projections of the objects' locations onto the images' planes,
  achieved by some key-point based matching process.
  
On each step (BA iteration), the environment receives $\lambda$ as an 
  action and weighs the GD and GN according to it, as is done in the classic LM scheme.
It then estimates the 3D locations of the objects and the camera poses, and  
projects these locations into 2D according to the estimated poses.
The stopping criterion is met when the estimation error is smaller than a certain threshold. 

The environment provides an observation (state {\color{blue_s}{$s$}}) and a reward ({\color{blue_r}{ $r$}}) on each iteration (step).
Let $z_{ij}$ be the ground truth pixel (match) in which key-point $j$ appeared in image $i$.
We model it as a noised projection of 3D-point $q_{j}$ on camera $c_{i}$ with a $w$ Gaussian projection noise, i.e $z_{ij} = Proj(c_{i},q_{j}) + w$.
Let $\hat{c_{i}},\hat{q_{j}}$ be the current iteration's estimated poses of the camera $i$ and location of 3D-point $j$ accordingly, and
let $\hat{z_{ij}}$ be the respective projection i.e $\hat{z_{ij}}=Proj(\hat{c_{i}},\hat{q_{j}})$. 
Let $\Delta z_{ij}$ be the difference between the ground truth projection and the estimated projection, i.e. $\Delta z_{ij} = z_{ij} - \hat{z_{ij}}$.
Let $C,Q$ be all the estimated camera poses and all the estimated 3D locations respectively.
The estimation error is set as the sum of $\Delta z_{ij}$, as follows:
\begin{equation} 
\begin{split}
&  Estimation \, error = \Sigma_{ci} ^ C \Sigma_{qj} ^ Q ||  {\mathbb{\Sigma}} ^{-1/2} \Delta z_{ij}||^2 \\
&  BA_{objective} = argmin_{CQ} [Estimation \, error],
\end{split}
\label{eq:state}
\end{equation}
where $\mathbb{\Sigma}$ is the covariance matrix, 
and the state ({\color{blue_s}{$s$}}) is set as a vector of the $5$ last consecutive errors, in order to enable the agent to learn the influence of the choice of $\lambda$ over a few iterations.
This forms the connection between the BA solving and the estimation error.

The reward ({\color{blue_r}{$r$}}) is set as the negative of the duration of each iteration (in seconds), apart from the convergence iteration (terminal state) where the reward is set as positive. 
In standard RL problems the agent is encouraged to maximize the sum of expected rewards: 

\begin{equation} 
\begin{split}
& E_{\pi}=\sum_{t=0}^{\infty}{r_{t}} \\
& r_{t} = - time_{BA iter_t} [seconds], 
\end{split}
\label{eq:reward loss}
\end{equation}
where $r_{t}$ is the reward at time step (iteration) $t$ received according to policy $\pi$. 
In our case, the agent is encouraged to minimize the overall processing time by reaching convergence, which indirectly minimizes the number of iterations.


\vspace{0.1in}
\noindent
{\bf Soft Actor Critic (SAC).} 
Our SAC framework consists of five networks that are updated according to the known actor-critic iterative optimization scheme:
\begin{enumerate}
    \item 
    an actor policy network that learns the actions. 
    As we aim at a learned $\lambda$ choosing, we chose the value of $\lambda$ as the one dimensional, real action;
    \item 
    two on-policy soft-critic
    networks, similar in structure, which evaluate the value function and differ in a time-delay;
    \item 
    an off-policy value network that evaluates the value function;
    \item
    a target network that converges the values predicted by the on-policy and off-policy networks into a single target value required for the actor-critic optimization.
\end{enumerate}

As the action represents the value of $\lambda$, it influences the optimization process directly.
Let $x$ be all the estimated camera's poses and 3D locations in the BA problem, $J$ be the Jacobin, $H$ be the Hessian, $\Delta z$ be a vector whose entries are $\Delta z_{ij}$ defined before, $\mathbb{\Sigma}$ be the covariance matrix
and $\lambda$ be the agent's action (damping factor).
The optimization step taken on each iteration to update $x$ is defined as:
\begin{equation} 
\begin{split}
& \Delta x = -\frac{1}{\lambda} J(x) ^{T} \mathbb{\Sigma} ^{-1} \Delta z \\
& {\text Optimization_{gradient}} = -(H+\lambda I) \Delta x.
\end{split}
\label{eq:action}
\end{equation}
This equation shows $\lambda$'s influence on the Jacobian $J$ and on the Hessian $H$, which impact the GD and GN, respectively.

\begin{figure}[tb]
\centering
\begin{tabular}{c}
\includegraphics[width=0.72\textwidth]{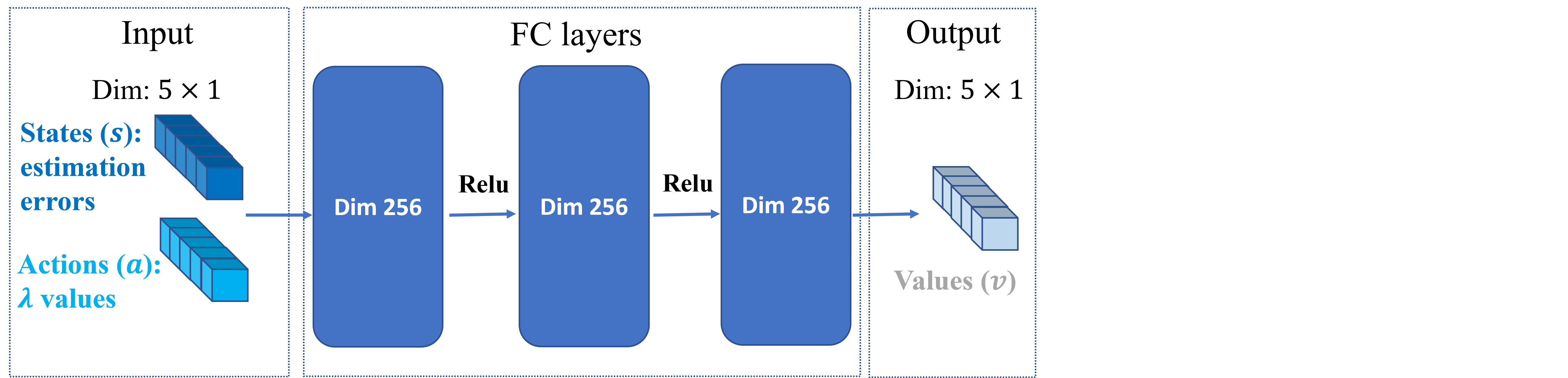}
\end{tabular}
\caption{
{\bf Soft-critic networks and Value network architecture.} The networks are similar in structure: $3$ Fully-Connected (FC) layers with Relu as the activation function, following ~\cite{rlalgorithms}'s implementation.
The value network receives the state vector (dim $5X1$) as input, while
both soft-critic networks networks receive the state and the action vectors (dim $5X2$) as input.
 }
 \label{fig:arch}
 \end{figure}

Following ~\cite{rlalgorithms}'s implementation, both soft-critic networks and value network have similar architecture: three FC layers (dim $256$), with Relu as the activation function, as seen in Fig.~\ref{fig:arch}.
The value network gets only the state as input, while the two soft-critic networks get both the state and the action.
The two differ in a $\tau$ iterations time difference (delay) only, and
if one of them reaches a sub-optimal evaluation the other is used instead on each iteration.
The target network gets the values (predictions) of the value network and the chosen critic
network and predicts a target (value) according to both on each iteration, and has a similar structure to that shown in Fig.~\ref{fig:arch}.
The policy network consists of four FC layers (dim $256$) with Relu as the activation function, gets the state as input and predicts the next action.
The loss functions of all networks are set to MSE.

\section{Experiments}
\label{sec:experiments}

\noindent
{\bf Datasets.} 
We ran our experiments on two real-life large datasets: KITTI~\cite{geiger2013vision} and BAL~\cite{agarwal2010bundle}, in which each scene may include tens of thousands of points.
While BAL provides both the camera poses and the 3D objects' key-points, KITTI provides only the camera poses with a series of images from which the key-points are extracted.

\vspace{0.1in}
\noindent
{\bf Results.} 
We compare our results to those of the classic BA approach with LM python implementation~\cite{more1978levenberg} and to two recent BA acceleration methods~\cite{zhou2020stochastic, huang2021deeplm}.
For each dataset we compare the results to the works that ran on that specific dataset.
The stopping condition threshold was set to 
 $1e^-6$
for all datasets.
As common, if the problem is not solved within $100$ iterations, it is considered as a failure. 
We use this definition to compare the success rate,
but note that {\em all} methods achieved $100\%$ success rate on both KITTI~\cite{geiger2013vision} and BAL~\cite{agarwal2010bundle}.
All time measurements are reported in seconds.
Each of the reported times includes both the approach's set-up time and its BA solving time.

Table~\ref{tbl: KITTI Average acceleration of the different methods} compares our results on the KITTI dataset. 
Our method required $1/5$ of the iterations to succeed in approximately $1/3$ of the time, with the same success rate.
Similar results are attained when comparing our results to other approaches on the BAL dataset, as shown in 
Table~\ref{tbl: BAL Average acceleration of the different methods}.

\begin{table}[tb]
\begin{center}
\begin{tabular}{|l|c|l|}
\hline
Method & $\#$Iterations & avg. time\\ 
\hline
 ~\cite{zhou2020stochastic} & $80$  & $287.4$ $[240]$  \\
 \hline
 Ours + ~\cite{zhou2020stochastic}  & $16$  & $110.0$ \\
 \hline
 Classic & $75$ & $340.0$ \\
 \hline
 Ours + classic &  $14$ & $100.2$ \\
 \hline
\end{tabular}
\vspace{0.1in}
\caption{
{\bf Average efficiency improvement
- KITTI.} 
Our approach accelerated the solving process by reducing number of iterations of other methods by a factor of $5$.
The total time diminished by a factor of $2.5$-$3.2$.
The time in the brackets is reported in~\cite{zhou2020stochastic}, whereas 
the time listed is for~\cite{zhou2020stochastic}'s implementation on our hardware.
}
\label{tbl: KITTI Average acceleration of the different methods}
\end{center}
\end{table}

\begin{table}[tb]
\begin{center}
\begin{tabular}{|l|c|l|}
\hline
Method &$\#$Iterations & avg. time \\ 
 \hline
 ~\cite{huang2021deeplm} & $15$ & $40.75$ $[25.2]$ \\
 \hline
 Ours + ~\cite{huang2021deeplm} & $5$ & $20.3$  \\
 \hline
 Ceres & $20$ & $46.7$ \\
 \hline
 Ceres + ours & $4$ & $20.75$ \\
 \hline
 Classic & $20$ & $110.0$ \\
 \hline
Ours + classic & $4$ & $62.04$ \\
 \hline
\end{tabular}
\vspace{0.1in}
\caption{
{\bf Average efficiency improvement - BAL.} 
Our approach accelerated solving process by reducing the number of iterations of other methods by a factor of $3$-$5$.
The total time diminished by a factor of $2$.
The time in the brackets is reported in~\cite{huang2021deeplm}, whereas 
the time listed is for~\cite{huang2021deeplm}'s implementation on our hardware.
}
\label{tbl: BAL Average acceleration of the different methods}
\end{center}
\end{table}


%
We compared the MSE between the final estimations and the ground truth on the BAL~\cite{agarwal2010bundle} dataset, when using~\cite{huang2021deeplm}'s method with our acceleration. 
We got similar results (difference $<$ 0.003) to those reported by~\cite{huang2021deeplm} on the same dataset.
This is not surprising as our method 
accelerates existing approaches and is not supposed to impact their accuracy.

\vspace{0.05in}
\noindent
{\bf Acceleration based on synthetic data}. 
Both KITTI~\cite{geiger2013vision} and BAL~\cite{agarwal2010bundle} contain large scenes,
and the scenes sizes directly affect the duration of each BA iteration.
Generally speaking, the bigger the scene the longer each iteration is.
Hence, utilizing either of these datasets for training requires a considerable amount of time.
We created a {\em synthetic random points} dataset to simulate smaller scenes, where the duration of each iteration is shorter, 
which highly reduces the overall training time.
This dataset was created by randomly selecting 3D locations (as points) and camera poses.
We created $10$ different trajectories, 
where each trajectory consisted of $10$ camera poses and $10$ locations, which differed between the different trajectories.
 We used these trajectories to train our agent, which was then tested on KITTI~\cite{geiger2013vision} and BAL~\cite{agarwal2010bundle}. 
\begin{table}[tb]
\begin{center}
\small
\begin{tabular}{|l|c|c|c|}
\hline
Method & Train data & Inference data & $\#$Iterations\\ 
\hline
 Ours + ~\cite{zhou2020stochastic} & Synthetic & KITTI & $16$ \\
 \hline
 Ours + classic & Synthetic & KITTI & $17$  \\
 \hline
 Ours + ~\cite{huang2021deeplm} & Synthetic & BAL & $5$  \\
 \hline
 Ours + classic & Synthetic & BAL & $5$ \\
 \hline
\end{tabular}
\vspace{0.1in}
\caption{
{\bf Average acceleration using synthetic data for training.} 
 Our method is able to accelerate the solving of~\cite{zhou2020stochastic, huang2021deeplm} and of the classic approach when trained on synthetic data.
The acceleration is similar to that achieved when using KITTI~\cite{geiger2013vision} and BAL~\cite{agarwal2010bundle} for training (see Tables~\ref{tbl: KITTI Average acceleration of the different methods} and~\ref{tbl: BAL Average acceleration of the different methods}), and reduced the number of required iterations by a factor of $3$-$5$.
}
\label{tbl:Average acceleration using random points for traning}
\end{center}
\end{table}

Table~\ref{tbl:Average acceleration using random points for traning} shows that using the synthetic data for training achieves similar acceleration to training on the original datasets.
Hence, our solution could be efficiently trained (time wise) on small synthetic scenes, and still successfully accelerate large real-life BA problems. 

\vspace{0.1in}
\noindent
{\bf Comparison to Ceres PCG on BAL.}
We compare Agarwal et al.’s~\cite{agarwal2010bundle} Ceres PCG C++ implementation to ~\cite{huang2021deeplm} accelerated by our method over the different BAL dataset's sub-problems. 
Table~\ref{tbl:BAL sub-problems} demonstrates that our method leads to a considerable acceleration.
\begin{table}[tb]
\begin{center}
\begin{tabular}{|l|c|c|c|}
\hline
Sub-problem & Ceres-CG & ~\cite{huang2021deeplm} & ~\cite{huang2021deeplm} + ours\\ 
\hline
Trafalgar & $65.1$ & $10.2 [1.81]$ & $5.6$ \\
 \hline
 Ladybug & $46.7$ & $15.35 [5.00]$ & $8.1$ \\
 \hline
 Dubrovnik & $320$ & $10.0 [3.14]$ & $4.1$\\
 \hline
 Venice & $1992$ & $60 [14.2]$ & $31.4$ \\
 \hline
 Final & $3897$ & $65 [24.4]$ & $33.7$ \\
 \hline
\end{tabular}
\vspace{0.1in}
\caption{
{\bf Acceleration of BAL's sub-problems.} 
Comparison of the different BAL sub-problems between Agarwal et al.’s~\cite{agarwal2010bundle} Ceres PCG C++ implementation to Huang et al.'s~\cite{huang2021deeplm} accelerated by our method. All times are reported in seconds. The times in the brackets are reported in~\cite{huang2021deeplm}, whereas 
the times listed are for ~\cite{huang2021deeplm}'s run time on our hardware.
}
\label{tbl:BAL sub-problems}
\end{center}
\end{table}

\vspace{0.1in}
\noindent
{\bf Implementation details.}
Following~\cite{rlalgorithms}'s PyTorch implementation, the SAC's training starts from a series of randomly chosen actions.
We used $500$ random choosings.
For the classic approach the initial value of $\lambda$ was set to $1/4$.
We used a continuous observation space that was set between $- \infty$ and $1000$
and a continuous action space that was set between $0$ and $\infty$.
The finish (convergence) bonus reward was set to $10$.
The time delay (\(\tau\)) between the soft-critic
networks was set to $5$ iterations.
The Adam optimizer was used for all networks.
We used a single NVIDIA RTX 3090 GPU for all experiments.

\section{Ablation Study}


\noindent
{\bf Comparison to a non-holistic learning scheme.}
Our key idea is to view the BA solving process in a holistic manner and to use RL
to learn the ideal value of $\lambda$.
We compared our RL approach to a non-holistic learning scheme that tries to learn the value of $\lambda$ 
 by minimizing the estimation error received on each iteration. 
For fair comparison,
we used a network with three FC layers, each at the size of $1280$, so it would have a similar number of parameters to that of the SAC framework.
Its input was set as three vectors, representing the last $5$ states, actions and rewards, so it would get the same information as the SAC framework.
We term it {\em Zero-net} as it aims to minimize the estimation error.

\begin{table}[tb]
\begin{center}
\begin{tabular}{|l|c|c|}
\hline
Method & $\#$Iterations\\
\hline
Classic & $20$  \\ 
\hline
Zero-Net + classic &  $8$ \\
\hline
Ours (holistic) + classic &  {$\mathbf 4$} \\
\hline
\end{tabular}
\vspace{0.1in}
\caption{
{\bf Comparison to a non-holistic learning scheme on BAL.} 
Our holistic RL based acceleration method is better than that of the non-holistic Zero-Net method.
}
\label{tbl: comparion to a non RL learning method on BAL}
\end{center}
\end{table}

We compared Zero-net and our method on the BAL dataset~\cite{agarwal2010bundle}.
Our method achieved superior results as seen in Table~\ref{tbl: comparion to a non RL learning method on BAL}, thus proving the importance of viewing the BA process in a holistic manner.

\begin{table}[tb]
\begin{center}
\begin{tabular}{|l|c|c|}
\hline
Method & Batch size & $\#$Iterations\\ 
\hline
 Classic & $-$ & $20$ \\
 \hline
 Ours + classic & $1$ & $7.5$ \\
 \hline
 Ours + classic & $5$ & {$\mathbf 4$} \\
 \hline
 Ours + classic & $10$ & $4.5$ \\
 \hline
 Ours + classic & $20$ & $5$ \\
 \hline
\end{tabular}
\vspace{0.1in}
\caption{
{\bf Acceleration with different state sizes on BAL.} 
The classic approach is accelerated by our method using different state sizes. $5$ achieves the best results.
}
\label{tbl:Acceleration using different batch sizes}
\end{center}
\end{table}

\vspace{0.05in}
\noindent
{\bf State size effect.} 
Each state represents $5$ consecutive estimation errors, to enable the agent to learn the influence $\lambda$'s value has over a few iterations.
To verify that $5$ iterations are sufficient, we trained our agent on the BAL dataset~\cite{agarwal2010bundle} with different state sizes.
The maximal size was set to $20$, as that is the number of iterations the classic approach required for BAL's solving.
Table~\ref{tbl:Acceleration using different batch sizes} verifies that $5$ is the optimal state size to enable the learning of $\lambda$'s value.

\vspace{0.05in}
\noindent
{\bf On the roles of the state and the reward.} 
This work proposes a method to reduce the overall time of the BA process, whilst maintaining the high accuracy that previous methods upheld.
We chose to represent the state as the estimation error and to represent the reward as the time.
Could these roles be reversed?
Could we use the time to represent the state and the estimation error to represent the reward?
\begin{table}[tb]
\begin{center}
\begin{tabular}{|l|c|c|}
\hline
Method & $\#$Iterations & Success rate\\ 
\hline
 Classic & $20$ & $100\%$ \\
 \hline
 Ours + classic & {$\mathbf 4$} & $100\%$ \\
 \hline
 Reversed + classic & $10$ & $70\%$ \\
 \hline
\end{tabular}
\vspace{0.1in}
\caption{
{\bf Acceleration with opposite state and reward roles.} 
The classic approach is accelerated by our agent and by a "reversed" agent, which gets each iteration's duration as a state and the estimation error as a reward.
The "reversed" agent accelerates the solving less than our method and reaches a lower success rate.
}
\label{tbl:Acceleration with state and reward reversed}
\end{center}
\end{table}
We ran such an experiment, attempting to accelerate the solving of the classic approach on the BAL dataset~\cite{agarwal2010bundle}, using the duration of each iteration (as a negative) as the state and the estimation error as the reward (with the same finishing bonus).
Table~\ref{tbl:Acceleration with state and reward reversed} compares our acceleration of the classic approach with that of the described "reversed" environment and agent. 
The "reversed" agent accelerated the solving less efficiently than our method, and achieved only $70\%$ success rate which is $30\%$ lower than {\em any} other approach.
This is probably since the "reversed" agent aims to minimize the overall error even at the expense of the solving's duration.

\vspace{0.05in}
\noindent
{\bf On the choice of the reward.} 
Apart from the convergence iteration (terminal state) where the reward is set as a positive convergence bonus, our reward was set as the duration of each iteration as a negative, which slightly differs between iterations.  
This raises a question of whether using a constant value as a negative reward ($-1$ in this experiment) instead of the duration would also suffice.
Another common reward format in RL is reward reduction, where the reward is set as $0$ in all iterations (states) but convergence (terminal state), and the positive finishing bonus is reduced (by $1\%$ of its value in this experiment) on every iteration.
In both cases, the longer it takes the agent to reach convergence, the smaller the 
sum
of its rewards would be.
This encourages the agent to reach convergence within as few iterations as possible.
\begin{table}[tb]
\begin{center}
\begin{tabular}{|l|c|}
\hline
Method & $\#$Iterations\\ 
\hline
 Classic & $20$ \\
 \hline
 Our reward & {$\mathbf 4$} \\
 \hline
 Constant negative reward & $5$ \\
 \hline
 Reward reduction & $4.2$ \\
 \hline
\end{tabular}
\vspace{0.1in}
\caption{
{\bf Acceleration with different rewards on BAL.} 
The classic approach is accelerated by our method using different rewards.
Using our reward produces better results than using a constant negative value as a reward and from using reward reduction.
}
\label{tbl:Different rewards types}
\end{center}
\end{table}
Table~\ref{tbl:Different rewards types} compares the acceleration results when using all three types of rewards on the BAL dataset~\cite{agarwal2010bundle}. 
The convergence bonus was set to $10$ in all cases.
All reward options successfully accelerated the solving, but our reward achieved the best results, 
probably due to the extra knowledge about the iterations duration that our reward provides.

\vspace{0.1in}
\noindent
{\bf Comparison to a constant (non-dynamic) scheduler.}
As discussed previously,  $\lambda$ weights between GD and GN, which impacts the number of iterations required to reach convergence.
There seems to be a pattern to the chosen $\lambda$ values.
In all the experiments descried in Tables~\ref{tbl: KITTI Average acceleration of the different methods} and~\ref{tbl: BAL Average acceleration of the different methods}, two small (close to $0$) values of $\lambda$ were chosen, followed by two bigger values. 

\begin{figure}[tb]
\centering
\begin{tabular}{c}
\includegraphics[width=0.45\textwidth]{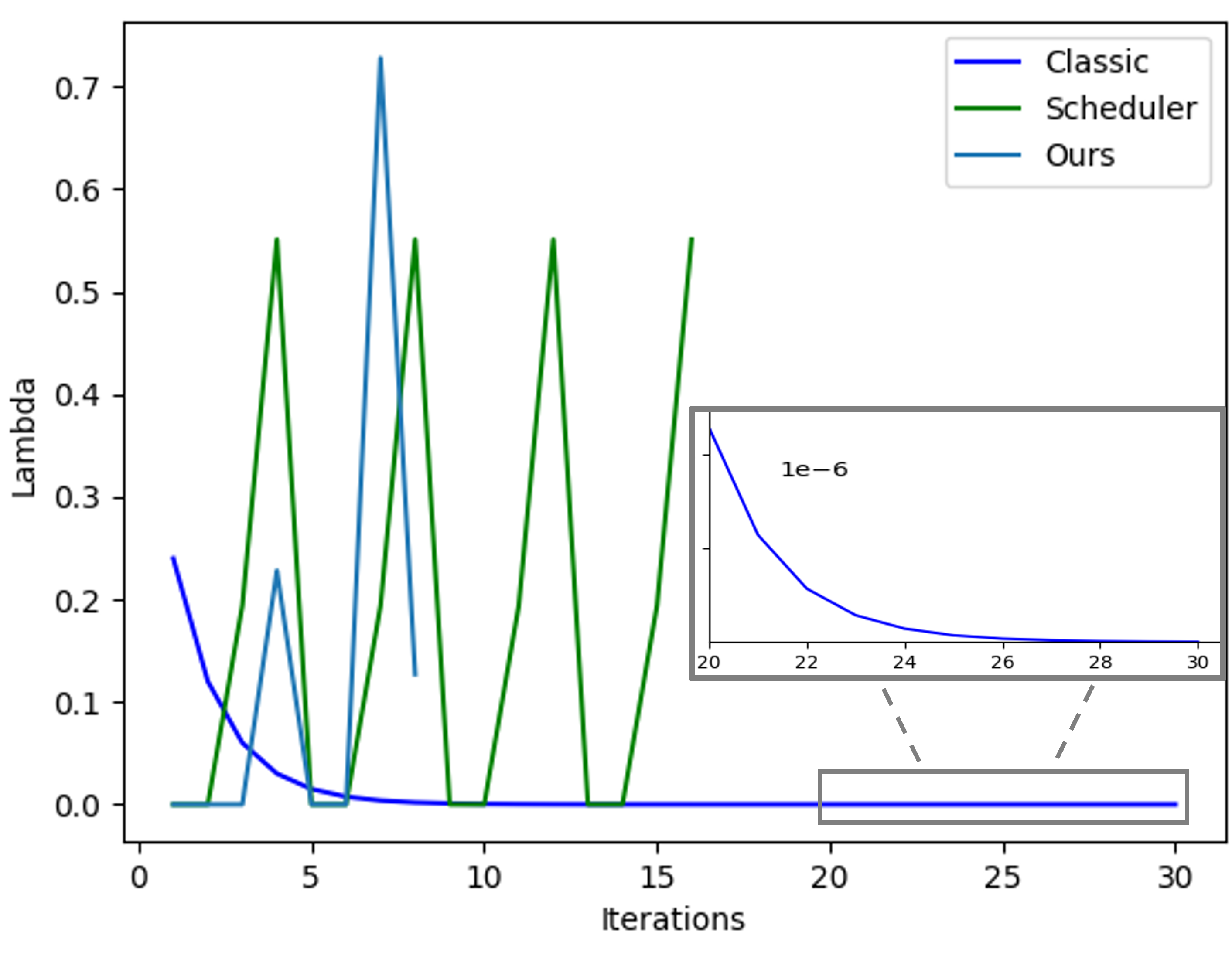}
\end{tabular}
\caption{
{\bf Chosen $\lambda$ values.} An example of the chosen $\lambda$ values in a single test case.
The classic approach requires $30$ iterations, the scheduler requires $16$ and our approach requires only $8$.
We zoom into the last $10$ iterations to demonstrate the decay of $\lambda$'s value to nearly $0$ by the classic approach.
 }
 \label{fig:Lmabda chart}
 \end{figure}

This raises a question of whether the optimal solution requires a dynamic value of $\lambda$, or
would scheduling a series of increasing and decreasing constant values of $\lambda$ suffice to reduce the number of iterations.
We tried to use the classic approach with such a scheduler on the BAL dataset~\cite{agarwal2010bundle}.
We ran the classic approach accelerated by our method on BAL
and extracted a constant series of four $\lambda$ values, that were taken as the average of the test scenes ([$1e-15$, $1e-15$, $0.194$, $0.551$]).
\begin{table}[tb]
\begin{center}
\begin{tabular}{|l|c|c|}
\hline
Method & $\#$Iterations & avg. time\\ 
\hline
 Classic & $20$ & $110.0$ \\
 \hline
 Constant scheduler & $12$ & $80.1$  \\
 \hline
 Classic + ours & {$\mathbf 4$} & {$\mathbf {62.04} $}  \\
 \hline
\end{tabular}
\vspace{0.1in}
\caption{
{\bf Comparison to a constant scheduler (non-dynamic method) on BAL.} 
When comparing our method's acceleration of the classic approach to that of a constant scheduler on the BAL dataset, our dynamic method achieves superior results.
The time is reported in seconds.
}
\label{tbl:comparison to aconstant scheduler}
\end{center}
\end{table}
This four long series was then used repetitively, as a constant (non-dynamic) scheduler to the classic approach.
Table~\ref{tbl:comparison to aconstant scheduler} shows that our approach needed only a $1/5$ of the iterations required by the classic approach to reach convergence and only a $1/3$ of the iterations required by the constant scheduler. 
This proves the importance of learning a {\em dynamic} value of $\lambda$ which {\em dynamically} weights GD and GN.
Figure~\ref{fig:Lmabda chart} shows an example of one of the described runs,
where the value of $\lambda$ changed by a constant factor by the classic approach. 
On the other hand, our dynamic method enabled $\lambda$'s value remain the same along three consecutive iterations which would not have been possible if a non-dynamic approach was used. 

\begin{figure*}[tb]
\centering
\begin{tabular}{cc}
\includegraphics[width=0.4\textwidth]{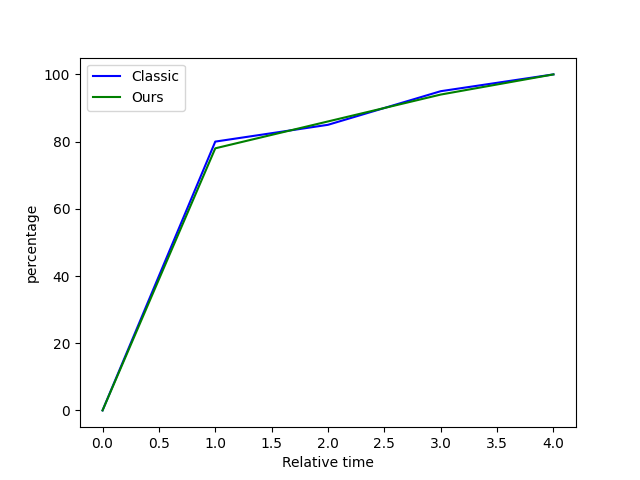} &
\includegraphics[width=0.4\textwidth]{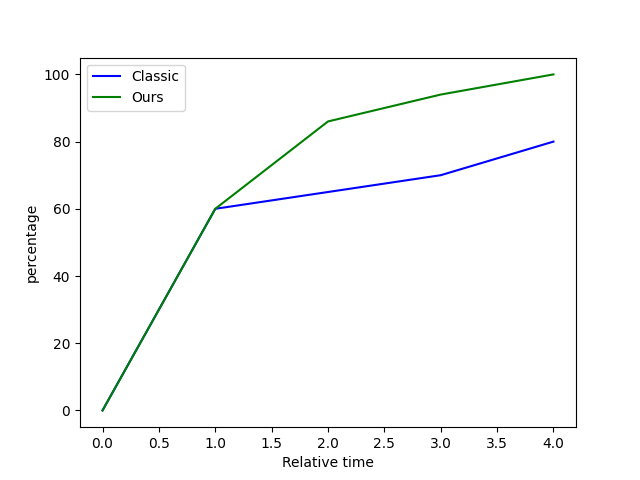}
\\
(a) tolerance $\tau = 0.1$  &
(b) tolerance $\tau = 0.001$
 \\
\label{fig:tolerence graphs}
\end{tabular}
\caption{
{\bf Performance profile on BAL.} (a) When a high tolerance ($\tau=0.1$) is used, the classic solution achieves a full solution very fast and our acceleration is not required. (b) When a more accurate ($\tau=0.001$) solution is required, our method accelerated the solving and reached a more accurate solution.
 }
\end{figure*}

\begin{figure}[tb]
\begin{center}
\includegraphics[width=0.7\linewidth]{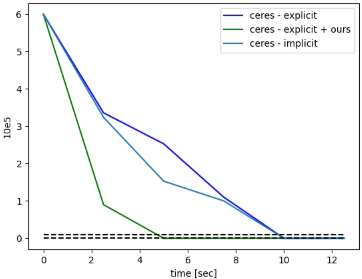}
\end{center}
   \caption{{Convergence plot of Ladybug-1197 from BAL.} The dashed lines are of tolerances $0.1$ and $0.001$, where explicit stands for PCG and implicit stands for Cholesky's method. }
\label{fig:1197}
\end{figure}

\vspace{0.1in}
\noindent
{\bf Convergence plot and performance profile on BAL.}
Each BA solver requires a different amount of time to solve a given BA problem. 
Let $\alpha$ be the time required by the fastest solver (e.g. minimal time required) to solve a given BA-problem.
Following ~\cite{weber2023power}'s formulation, performance profiles are used to compare how much of a given BA problem (in percentage) is solved by different solvers, while using $\alpha$ as the relative time. 
Fig~\ref{fig:tolerence graphs} demonstrates that when a very high threshold ($\tau = 0.1$ tolerance) is used, the classic approach reaches a solution within very little time and requires no acceleration.
However, when a more accurate solving is required ($\tau = 0.001$ tolerance), our acceleration enables reaching higher accuracy within less time.


Convergence plots are used to evaluate the performance of different solvers without using relative time. An example of a convergence plot of scene 1197 from BAL's Ladybug sub-problem is shown in Fig.~\ref{fig:1197}

\vspace{0.05in}
\noindent
{\bf Limitations.}
When considering small BA problems or high tolerance ($\tau = 0.1$ threshold for instance), which are commonly solved in a few iterations by the classic approach, we cannot improve them to the same extent as bigger BA problems.
For instance, when $5$ random points and $5$ random camera poses are used, the classic approach reaches convergence within $5$ iterations on average while our method required $4$ iterations on average.
Furthermore, our method's overall solving time was $1/2$ a second longer than that of the classic approach in this case, due to the agent's inference time.
Therefore, when considering small or inaccurate solving of BA problems our method is less effective
, and may even perform worse on extremely small problems.



\section{Conclusion}
Localization and mapping are key problems in many real time applications, that 
 are commonly solved using the iterative Bundle Adjustment (BA) process. 
On each BA iteration, a system of non-linear equations is solved using two optimization methods: Gradient Descend (GD) and Gauss-Newton (GN), each better suited for different parts of the solving.
In the classic approach, these two methods are weighted by a damping factor, $\lambda$, 
that may change by one of two {\em constant} factors between consecutive iterations.
This may prevent the classic approach from efficiently weighting between GD and GN 
which might result in many iterations.

Our key idea is therefore to learn a dynamic value of $\lambda$ in order to reduce the sheer number of iterations required to reach convergence by efficiently weighting GD and GN.
This is not trivial as 
the solving needs to be considered as a whole in order to learn from it.
Hence, differently from past approaches, we propose to view the BA process in a holistic manner as a game.
We use a Reinforcement Learning (RL) based method to learn $\lambda$, as it can handle sparse and delayed rewards like the BA's convergence and views the BA process as a whole. 
We use the Soft Actor Critic (SAC) RL framework as it is stable, well suited for continuous state and action spaces, and for its extensive exploration. 

We set an environment that solves the non-linear equations system, and an agent who's action determines the value of $\lambda$.
The reward is set as negative in all iterations, apart from the convergence iteration where the reward serves as a positive convergence bonus.
As the agent aims at maximizing the sum of expected rewards, it is encouraged to solve the problem within as few iterations as possible.
This is the key to our method's time reduction.

Our RL based solving approach is shown to reduce the number of iterations required to reach convergence by a factor of $3-5$ on both known KITTI~\cite{geiger2013vision} and BAL~\cite{agarwal2010bundle} benchmarks.
Our reduction could be especially meaningful in real life scenarios that may require much time to solve due to their large size.
Our agent may also be trained on small synthetic scenes, which is highly time-efficient,  
and still accelerate the solving of bigger real-life scenarios. 
 Moreover, our approach may be added to previous acceleration methods that focus on reducing the time of each iteration, as demonstrated on
two different methods.
However, hardware based BA optimization methods such as~\cite{weber2023power, ortiz2020bundle, Demmel_2021_CVPR} may benefit less from our acceleration.

For future work we would like to further optimize the agent's architecture to the task at hand, and extend this method to other optimization problems, such as SfM.

\vspace{0.05in}
\noindent
{\bf Acknowledgements.}
We gratefully acknowledge the support of the Israel Science Foundation (ISF) 2329/22.
We thank Evyatar Bluzer, Maurizio Monge, True Price, Ruibin Ma, Jonathan Zeltser, Omri Carmi and Jan-Michael Frahm for their valuable comments and assistance.


\newpage
{\small
\bibliographystyle{ieee_fullname}
\bibliography{egbib}
}

\end{document}